# Automatic segmentation of colorectal liver metastases for ultrasound-based navigated resection

Tiziano Natali[1,2,*], Karin A. Olthof[1,2], Niels F.M. Kok[1], Koert F.D. Kuhlmann[1], Theo J.M. Ruers[1,2], Matteo Fusaglia[1]

[1] Department of Surgical Oncology, Netherlands Cancer Institute, Plesmanlaan 121, 1066CX, Amsterdam, The Netherlands

[2] Faculty of Science and Technology (TNW), Nanobiophysics Group (NBP), University of Twente, Drienerlolaan 5, 7522 NB, Enschede, The Netherlands

**\* Corresponding author:** t.natali@nki.nl, +31 20 512 1490

**Fuding:** This research received no external funding.

**Category:** Original article

**Conflicts of Interest:** The authors declare no conflict of interest

**Data Availability Statement:** The datasets generated during and/or analyzed during the current study are not publicly available.

**Ethical Statement:** All procedures performed in this study involving human participants were conducted in accordance with the ethical standards of the institutional and national research committee.

# Abstract

**Introduction:** Accurate intraoperative delineation of colorectal liver metastases (CRLM) is crucial for achieving negative resection margins but remains challenging using intraoperative ultrasound (iUS) due to low contrast, noise, and operator dependency. Automated segmentation could enhance precision and efficiency in ultrasound-based navigation workflows.

**Methods:** Eighty-five tracked 3D iUS volumes from 85 CRLM patients were used to train and evaluate a 3D U-Net implemented via the nnU-Net framework. Two variants were compared: one trained on full iUS volumes and another on cropped regions around tumors. Segmentation accuracy was assessed using Dice Similarity Coefficient (DSC), Hausdorff Distance (HDist.), and Relative Volume Difference (RVD) on retrospective and prospective datasets. The workflow was integrated into 3D Slicer for real-time intraoperative use.

**Results:** The cropped-volume model significantly outperformed the full-volume model across all metrics (AUC-ROC = 0.898 vs 0.718). It achieved median DSC = 0.74, recall = 0.79, and HDist. = 17.1 mm—comparable to semi-automatic segmentation but with ~4× faster execution (≈ 1 min). Prospective intraoperative testing confirmed robust and consistent performance, with clinically acceptable accuracy for real-time surgical guidance.

**Conclusion:** Automatic 3D segmentation of CRLM in iUS using a cropped 3D U-Net provides reliable, near real-time results with minimal operator input. The method enables efficient, registration-free ultrasound-based navigation for hepatic surgery, approaching expert-level accuracy while substantially reducing manual workload and procedure time.

# 1  Introduction

Surgical resection remains the preferred treatment for colorectal liver metastases (CRLM), offering the longest prognosis [1]. The goal of surgery is to achieve complete tumor removal (i.e. a negative resection margin), as positive margins are strongly associated with local recurrence and poor prognosis [2]. Yet, intraoperative delineation of tumor boundaries remains a significant challenge, particularly when relying solely on intraoperative ultrasound (iUS). Ultrasound is a widely adopted modality because of its high availability, provides real-time imaging, and safety profile [3]. However, it is also limited by operator dependency [4], the absence continuous feedback during resection, and the cognitive burden it places on surgeons, who must mentally reconstruct three-dimensional anatomy from rapidly changing 2D images. In addition, during resection, gas bubbles generated by electrocautery interfere with the ultrasound signal, further complicating accurate margin assessment.

Image guidance systems have been developed to provide real-time navigation together with advanced visualization methods. IUS has been integrated in several guidance systems [5-7], which have solved some of the inherent limitations of iUS as imaging modality by integrating real-time tracking of surgical instruments. These studies managed to merge 3D models generated by segmentations from pre-operative MRI or CT scans in the surgical scene, thus adding spatial knowledge of the position of the iUS with respect to such models.

Although these systems managed to integrate volumetric anatomical information in the surgical scene, they need to perform a so-called registration step, which is required to set the spatial relation between the intraoperative scene with the preoperative models. This step is mostly performed by selecting anatomical landmarks in the pre-operative models, which are then found back in the intraoperative US image. This can result to be time-consuming and prone to errors due to organ movement, deformations in addition to operator dependency and inconsistent landmark identification [5]. Recent studies [8, 9] proposed surgical navigation workflows that skip the registration step. In these registration-free navigation workflows, the surgery is guided by leveraging the anatomical information retrieved only from the tracked iUS probes, by manually or semi-automatically segmenting the relevant anatomies present in the tomographic reconstruction, which is generated from tracked iUS sweeps over the area of interest.

Manual delineation of CRLM in iUS is both time-consuming and operator-dependent due to the heterogeneous appearance of liver parenchyma and metastases. Lesion margins are often poorly defined because of low contrast between tumor and surrounding tissue, continuously changing images, acoustic shadowing, and the presence of speckle noise inherent to US imaging. In addition, CRLM can display variable echogenicity (appearing hyperechoic, hypoechoic, or isoechoic relative to the parenchyma) making manual segmentation particularly challenging. Consequently, intraoperative resection planning can be delayed and variability can be introduced across clinicians. Automating this step has the potential to significantly improve the speed, consistency, and reproducibility of surgical navigation. To address the challenges of intraoperative tumor localization, we propose an automated tumor segmentation algorithm providing accurate, intraoperative delineation of CRLM to support surgical decision-making and margin assessment.

In the latest years, thanks to the recent developments in Deep Learning (DL) algorithms, automated tumor segmentation in ultrasound has been shown to be achievable with good accuracy and low algorithmic running time, showing promise in overcoming the limitations of manual interpretation [4, 10]. Recent studies have demonstrated that supervised DL models, such as U-Net and its variants, can achieve segmentation accuracy in iUS images of liver metastases comparable to inter-observer segmentations [4, 11, 12].

Most prior research on US tumor segmentation has relied on 2D frame-wise models, which predict each image independently and often suffer from poor temporal consistency across sequences. One representative work is by Leclerc *et al.* [13], who organized the CAMUS challenge for cardiac US and benchmarked several deep learning architectures for 2D segmentation, demonstrating the power of CNNs for US but also revealing significant instability across frames, particularly when motion or acoustic shadowing was present. These studies established strong baselines but confirmed that 2D models, while accurate per frame, are prone to temporal inconsistency and lead to unstable volumetric reconstructions. Similar results have been presented in the work by Natali et al. [4], which introduced a 2D model for hepatic tumor segmentation in iUS that achieved promising segmentation results in single images of clear lesions, but shown inconsistent results when tested on longer US sequences.

To address this, recent work has explored "2.5D" approaches, which incorporate information from neighboring frames without the full computational cost of 3D models. Zhou et al. [14] proposed a 2.5D strategy for cardiac MRI and echocardiography segmentation by stacking adjacent slices as input channels, achieving smoother contours and improved robustness to noise compared to pure 2D methods. In parallel, Smistad *et al.* [15] developed a real-time CNN with LSTM layers for cardiac US segmentation which managed to achieve more stable segmentations across consequent frames and outperforming standard post-processing methods. Both studies demonstrated that leveraging short-term temporal context yields more stable and anatomically consistent predictions.

Despite these advances, significant challenges remain before such methods can be translated to hepatic tumor segmentation from intraoperative US. First, the majority of prior work has been developed on cardiac US, where structures are relatively large, repetitive, and surrounded by consistent anatomy. Liver tumors on the other hand, are heterogeneous in echogenicity and appearance, and surrounding anatomies can change drastically in the span of few frames while scanning. Second, the problem of small lesion size (with more than half of colorectal liver metastases measuring <20 mm) exacerbates instability, as minor segmentation errors across frames can obscure or erase lesions entirely in volumetric reconstructions. Finally, intraoperative US is further affected by strong speckle noise, rapid probe motion, and operator dependency, which collectively make temporal consistency even harder to achieve.

To address these limitations, we propose a framework that leverages the volumetric context of tracked iUS data to train a 3D U-Net. High-quality tomographic reconstructions from iUS can be achieved in surgical navigation setups like the one presented in the work by Smit et al. [16], and these reconstructions can be fed to a DL model for automatic tumor segmentation. Unlike standard 2D CNNs, the 3D U-Net architecture can capture spatial dependencies and anatomical continuity across sequences of US frames, leading to improved boundary delineation and reduced false positives in the heterogeneous liver parenchyma. We hypothesize that focusing on localized regions can further improve segmentation performance, increasing tumor detection rate and decreasing manual workload for the technical clinicians, finally reducing overall operation time.

The presented method is meant to be introduced in a surgical navigation workflow similar to that presented in the ex-vivo work by Paolucci et al. [8], where tumor margins are delineated on a central iUS frame, which is then used to generate a spherical approximation. This assumption limits the performance of the model especially on large and irregularly shaped tumors Olthof et al. [9] implemented and tested in-vivo a similar workflow, and improved the tumor margin delineation by semi-automatically segmenting tumors in the iUS reconstructed sweeps using a seeded region growing algorithm. This resulted in more accurate tumor margins, at the cost of extra operation time. Our work aims at further improving this workflow by introducing an automated DL segmentation algorithm, achieving segmentation results similar to semi-automatic approaches and at the same time drastically reducing time required for the tumor margin delineation. The key contributions of this study are summarized as:

- The introduction of an automated 3D segmentation approach for CRLM tumors in intraoperative ultrasound;
- A comparative analysis of models trained for segmenting on full iUS volumes vs. cropped ones;
- A prospective evaluation in surgical cases, which indicated that the model is suitable for use in clinical workflows.

These findings suggest that accurate tumor segmentation can be achieved using intraoperative ultrasound alone, without reliance on preoperative imaging or manual annotations. This supports the rationale of developing ultrasound-based navigation systems that are less dependent on complex registration workflows, and may enhance the efficiency and accuracy of tumor localization during liver surgery.

## 2 Methods

### 2.1 Data Acquisition

The data used in this study consisted in 85 3D iUS volumes from 85 patients, scheduled for open resection of CRLMs. The volumes are tomographic reconstructions generated from sequences of 2D iUS frames (sweeps or videos) coupled with their respective positions and rotations. Spatial information was acquired using an NDI® Aurora electromagnetic (EM) tracking system (Northern Digital, Waterloo, Canada). iUS sweeps were acquired using an EM-tracked T-shaped convex ultrasound transducer (I14C5T model, BK Medical ApS, Peabody, MA), operating at a frequency between 5 to 14 MHz and a focusing distance between 2.5 to 6.5 cm. Acquired iUS sequences ranged in number of images between 38 to 95, with resulting reconstructions volumes between 160 and 355 $mL$. iUS volumes used in this study were sourced from two different studies, as summarized in Tab. 1. Dataset *A* is retrospective and includes 55 US sweeps from 60 patients, which were a part of a larger clinical study (IRBd20-091). Dataset *A* contains iUS volumes imaging a single CLRM, characterized by hyper- ipo- and mixed-echoic appearance and visible margins. Iso- and nearly-iso-echoic CLRM were excluded to avoid hindering model performances, as this was suggested in a previous study [4]. Dataset *A* was used during the training phase of the model implemented in the proposed workflow, and was split in Train and Validation sets (see Tab. 1), of 50 and 10 volumes respectively. Dataset *B* was accrued as part of a clinical study aimed at implementing a surgical navigation workflow for the resection of hepatic lesions using intraoperative ultrasound only [9] (NL80634.031.22). The dataset consists of 25 volumes from 25 patients and was used for testing the developed models. 20 volumes were used for retrospective testing (Dataset $B_{retro}$), while automatic segmentation of the other 5 was prospectively tested during the procedures ($B_{pro}$).

### 2.2 Ground Truth Delineation

Ground truth delineation of all volumes was performed retrospectively by a single experienced medical technician (5 years of experience in hepatic iUS images assessment). During tumor delineation, the expert had access to patient reports and diagnostic scans. Delineations were performed using 3D Slicer [17] version 5.6.2 as software. The output of the manual delineation was a set of 85 binary masks.

### 2.3 Model

We used a 3D U-Net architecture implemented via the nnU-Net framework. The version of nnU-Net was downloaded in July 2024 and executed using PyTorch 2.6 and Python 3.10. All training was conducted on a single NVIDIA GeForce GTX 1080 Ti GPU. The model was trained using the default configuration parameters. More specifically, training was performed without any data augmentation, using the Adam optimizer with an initial learning rate of 0.01, employing a polynomial learning rate schedule with warmup iterations as defined in the nnU-Net framework. The loss function was a combination of DSC and cross-entropy, balanced equally to optimize both region overlap and voxel-level classification accuracy. The model input consisted of a single-channel 3D iUS volume, and the output was a binary mask indicating the predicted tumor location.

Two variants of the model were trained using the training dataset described in Tab. 1 and the corresponding manual annotations. The first model ($Model_{crop}$) was trained on volumes cropped around the tumor, with an additional 1 cm margin in all directions to preserve contextual information while reducing irrelevant background. The second model ($Model_{full}$) was trained on the entire 3D iUS volumes, using the full spatial context available during acquisition. In both cases, input and output volumes were resampled and normalized according to nnU-Net's specifications.

To determine the optimal decision threshold for converting the model's probabilistic outputs into binary segmentations, Precision-Recall (PR) analysis was performed on the validation set. Each voxel within the ultrasound volumes was treated as an individual prediction, and the threshold corresponding to the maximum F1-score was selected. The selected threshold was determined exclusively on the Validation set (Tab. 1) and then applied without further tuning during testing phase, ensuring an unbiased evaluation of model generalization capabilities. Receiver operating characteristics curves and relative AUCs were also computed to further detail the models' performances.

### 2.4 Intraoperative Implementation

To facilitate real-time clinical usability, the proposed segmentation workflow was implemented as a custom module within the 3D Slicer platform [11]. The workflow, illustrated in Fig. 1, allows the user to interactively acquire and process iUS volumes with minimal disruption to the surgical workflow.

During surgery, the user initiates the acquisition by recording a sequence of tracked 2D iUS frames which are subsequently reconstructed into a 3D volume. Once reconstruction is completed, the user defines a region of interest by manually placing a bounding box around the tumor. This cropped volume is then automatically passed to the segmentation model which returned a volume with voxel-wise tumor prediction.

The resulting segmentation, rendered as a 3D model in the navigation scene, is overlaid on the iUS video stream and visualized with respect to the other tracked surgical instruments. This enables immediate visual feedback to the surgeon during image-guided resection. When necessary, adjustments to the segmentation could be performed manually. These corrections were included in the overall time required for the segmentation process.

### 2.5 Evaluation

The performances of the proposed segmentation models were evaluated through both spatial overlap metrics and classification-based performance measures, using two distinct datasets ($B_{retro}$ and $B_{pro}$). For the evaluation on $B_{retro}$, the primary metric used to quantify segmentation accuracy was the 3D Dice Similarity Coefficient (DSC), calculated between the model predictions and the manually annotated tumor masks. DSC scores were also computed for semi-automatic segmentations to serve as a clinical reference and establish a benchmark for intraoperative performance.

To complement the DSC and better characterize the geometric fidelity of the segmentations, the 95th percentile Hausdorff Distance (HDist.) was also computed. This metric provides insight into boundary agreement and the presence of large outlier errors, which are especially critical in surgical applications where accurate margin delineation is essential. Relative Volume Difference (RVD) was also calculated to assess agreement in the tumor's size estimation and to evaluate the model's ability to preserve volumetric fidelity across tumors of varying sizes.

Additionally, lesion-level sensitivity and specificity were computed to assess the model's detection capabilities. In this analysis, a lesion was considered correctly detected if the DSC between the predicted and ground truth segmentation exceeded 50%.

Segmentation accuracy on $B_{pro}$ was assessed postoperatively by comparing the model-generated segmentations with both manual and semi-automatic annotations. The same metrics were applied, enabling consistent quantitative comparison between the fully automatic and the expert-guided segmentation strategies within clinical settings.

To evaluate practical feasibility, runtime performance was also assessed by recording the time required to generate automatic segmentations and comparing it with the time needed to complete semi-automatic labeling. Operation times for both methods were measured retrospectively from screen recordings of the navigation software. This allowed for a consistent and unbiased assessment of the time required for each segmentation strategy under real-world usage conditions.

To assess statistical significance, Wilcoxon's signed-rank test was used to compare paired results. An initial significance threshold was set at $\alpha = 0.05$. However, to account for multiple comparisons, the Bonferroni correction was applied, adjusting the threshold $\alpha_{adj} = \frac{\alpha}{n_{comparisons}} = \frac{0.05}{3} = 0.017$. Accordingly, results were considered statistically significant only when $p < \alpha_{adj}$.

# 3 Results

## 3.1 Retrospective Results

$Model_{crop}$ significantly outperformed $Model_{full}$ across all evaluation metrics. Specifically, $Model_{crop}$ achieved an area under the ROC curve (AUC-ROC) of 0.898, compared to 0.718 for the Modelfull, indicating superior sensitivity-specificity trade-offs. Likewise, the area under the PR curve (AUC-PR), was substantially higher for $Model_{crop}$ (0.755) relative to $Model_{full}$ (0.363). The optimal classification threshold, highlighted on the PR curve, resulted to generate masks with accuracy values of 0.66 for precision and 0.70 for recall, respectively. This threshold was subsequently used for binary mask generation and performance evaluation on the set B (Tab. 1).

Figure 3 presents a comparative overview of segmentation performance and runtime across all evaluated methods. The semi-automatic approach consistently shows the highest scores in DSC, precision, and recall. These values reflect high agreement with ground truth annotations and minimal variability across cases. However, this superior accuracy comes at the cost of a significantly higher runtime, with median execution times of approximately 230 seconds, $\approx 250\,\%$ slower than the automated methods. $Model_{crop}$ achieved performances second only to the semi-automated method across all segmentation metrics, with median values of 0.74, 0.76, and 0.79, respectively and compact distributions. However, statistical analysis indicates that these differences are not significant, with $p$ of 0.054, 0.024, 0.58, 0.14, 0.06 for DSC, precision, recall, VRD and HD, respectively. On the other hand, $p$ of 0.010 for precision, 0.0024 for VRD and $p < 0.001$ for DSC, recall and HD confirm that $Model_{crop}$ significantly outperforms the Modelfull across all metrics. This suggests that localization-based inputs enhance both accuracy and reliability. $Model_{crop}$ also achieves near real-time processing, with execution times typically under one minute, making it well-suited for intraoperative use. $Model_{full}$ achieved the poorest overall performance, presenting substantial improvement when removing the missed lesions (i.e., $DSC = 0$), but still underperforming with respect to the other methods. Importantly, this variant was not included in the statistical analysis, as its selective nature introduces bias and does not represent a full evaluation of model behavior. The HDist. provides further insight into boundary agreement. The semi-automatic approach and $Model_{crop}$ both achieve relatively low HDist. values, with medians of 12.8 mm and 17.1 mm, respectively. These suggest that, even in cases where volumetric agreement (e.g., DSC) may slightly drop, the predicted boundaries remain within clinically acceptable distances from the ground truth. In contrast, $Model_{full}$ displays the highest HDist. values, underscoring poor localization performance and higher segmentation errors. Finally, the trends for VRD mirror those observed in DSC and recall: the semi-automatic approach and $Model_{crop}$ maintain lower and more stable values, indicating better agreement in segmented lesion size. Modelfull again shows large deviations, consistent with its high rate of missed detections and over-/under-segmentations. Overall, Fig. 3 clearly shows that that $Model_{crop}$ provides the most favorable balance between segmentation accuracy and operational efficiency, achieving statistically significant improvements over Modelfull while approaching the accuracy of manual methods at a fraction of the time, only requiring $\approx 1$ minute per case and minimal operator interaction.

To further illustrate the behavior of the automatic segmentation method, Fig. 5 presents qualitative examples including a worst-case, two representative average cases, and a best-case scenario. The worst-case example shows a lesion with low contrast and complex background anatomy, where the model failed to produce a meaningful segmentation, resulting in low DSC and high Hausdorff values. In contrast, the two average cases reflect stable and anatomically plausible segmentations with minor boundary discrepancies, correlating well with median quantitative metrics. Finally, the best-case scenario exhibits near-perfect alignment with the ground truth, demonstrating the model's capacity for accurate delineation under favorable imaging conditions. These examples underscore the practical variability of model outputs and support the statistical findings from the quantitative evaluation.

## 3.2 Prospective Results

Patients prospectively included in dataset $B_{pro}$ resulted to have median age of 55 years with range of 48 and 67. Median tumor volume size, computed from the manual segmentations performed over the reconstructed iUS volumes (Sec. 2.2) resulted to be of 2.68 mL within range of 0.56 and 35.6 $mL$. Median tumor diameter was of 15.9 mm within range of 10.1 and 40.6 mm.

Figure 4 presents violin plots summarizing the segmentation performance and runtime comparisons on the dataset $B_{pro}$, of $Model_{crop}$ against the semi-automatic approach. The results confirm trends observed in $B_{retro}$, with $Model_{crop}$ maintaining similar performance across all metrics. $Model_{crop}$ maintained segmentation accuracy comparable to the semi-automatic approach across all primary metrics, confirming the robustness of the method during intraoperative use. A slight increase in execution time was observed for $Model_{crop}$, while the semi-automatic approach required marginally less time than in the retrospective analysis. This inversion is likely attributable to the change in use context, where, unlike in the retrospective evaluation, the intraoperative tumor segmentations were generated by $Model_{crop}$, whereas the semi-automatic segmentations were performed postoperatively. Despite this shift, statistical analysis confirmed that the differences in segmentation performance remained consistent with those observed retrospectively, further supporting the applicability of the automatic method in clinical settings.

## 4 Discussion and Conclusions

This study demonstrates that automated segmentation of colorectal liver metastases in intraoperative ultrasound can achieve clinically meaningful accuracy with minimal operator intervention. Among the evaluated strategies, the model trained on cropped volumes provided the best balance between segmentation performance and runtime efficiency. It significantly outperformed $Model_{full}$ across all key metrics and approached the performance of the semi-automatic approach (i.e., manually seeded region-growing algorithm), while requiring only a fraction of the time to execute. $Model_{crop}$ achieved a median Dice of 0.74 and recall of 0.79, with a corresponding Hausdorff Distance (HDist) of 17.1 mm. These values fall within clinically acceptable boundaries for surgical planning and suggest the model's outputs are well-aligned with expert annotations in most scenarios. Although semi-automatic segmentations retained a marginal advantage in absolute accuracy, statistical testing indicated that the differences between the two methods were not significant for most metrics, underscoring the reliability of the automatic approach.

In contrast, $Model_{full}$ consistently underperformed. Its wider metric distributions, lower Dice (median 0.03), and recall (median 0.01), alongside higher HDist values (often > 30 mm), indicate that processing the entire liver volume lowers the model's focus and hinders its ability to localize lesions accurately. Even when excluding zeros from the performance of $Model_{full}$, segmentation results scores remained lower than those of the cropped approach.

The automatic cropped model's segmentation quality was further supported by qualitative analyses. In favorable conditions, with clear lesion margins and good contrast, segmentation closely matched ground truth, presenting minimal boundary discrepancies. Average cases showed robust and plausible predictions, while poor performance was largely confined to rare low-contrast, anatomically complex scenarios. Notably, the model maintained acceptable precision and boundary error even in these challenging cases, suggesting that its failure modes are predictable and interpretable.

Importantly, the model achieves low inference times, requiring median operation runtime under one minute per volume, resulting to be approximately 4× faster than the semiautomatic approach. By significantly reducing the time required for segmentation without compromising reliability, the model supports integration into navigated workflows for real-time margin assessment and surgical planning.

Although the promising results shown in this study, there are some limitations that should be considered when interpreting the results and planning future work. The dataset used to train and evaluate the segmentation model was collected at a single clinical site using one specific ultrasound scanner and acquisition protocol. While this ensures consistency within the study, it may limit the model's ability to perform reliably when applied to data from other hospitals or with different imaging setups. Differences in scanner hardware, patient populations, and how the ultrasound probe is handled during acquisition can all influence image appearance and affect the model performance. To improve generalizability, future work should consider assembling multi-center datasets that reflect a broader range of US scanners, including images from laparoscopic or robotic iUS and percutaneous US. In situations where data sharing is restricted, federated learning frameworks offer a promising alternative, allowing models to be trained across institutions without transferring patient data, while maintaining strong performance across varied imaging conditions [18, 19]. The dataset deliberately excluded iso- and nearly isoechoic lesions due to their inherently poor visibility in conventional B-mode intraoperative ultrasound, a limitation of the physical imaging modality rather than related to data quality. Consequently, the model is currently applicable only to lesions with sufficient contrast in B-mode imaging, and does not generalize to those that resemble the parenchyma. To address this gap, future research should consider integrating contrast-enhanced ultrasound (CEUS), which can highlight vascularized lesions via microbubble perfusion. This technique has shown promise in renal tumor segmentation with CNNs [20]. Additionally, shear-wave elastography (SWE) provides quantitative tissue stiffness maps that can reveal lesions invisible in B-mode, and recent deep learning frameworks have facilitated lesion characterization and segmentation based on elastography signals with promising accuracy [21]. Combining B-mode with CEUS and or SWE in a multi-modal framework would potentially extend the method's applicability to clinically relevant but acoustically hidden lesions. Although this study demonstrated the feasibility of developing an automatic tumor segmentation model with clinically acceptable accuracy, the model's performances could be further improved. Specifically, while ground truth segmentations were produced by a small group of expert annotators without formal modeling of inter- or intra-observer variability, prior research has shown that accounting for annotator disagreement can enhance training reliability, particularly in subjective tasks like tumor boundary delineation. Deep-learning-based methods that incorporate annotator-specific uncertainty or aggregate multiple labels into probabilistic consensus masks have demonstrated improved segmentation performance under noisy-label conditions [22]. Integrating such techniques in future work could therefore increase the robustness and reproducibility of the training process, has this has been proven in other fields [22]. Incorporating such annotation uncertainty frameworks or consensus labeling procedures would strengthen label quality and model trustworthiness. The cropping strategy, while beneficial for runtime and accuracy, inherently sacrifices broader anatomical context. Hybrid architectures, such as dual-branch global-local fusion networks, have been shown to capture fine texture details while preserving global structure, yielding improved segmentation in organs and lesions alike [23, 24].

Taken together, these findings support the integration of automatic segmentation into iUS-based surgical navigation. When paired with the appropriate clinical safeguards and visual quality control, the cropped-input deep learning model offers a compelling step toward fully automated, efficient, and accurate intraoperative tumor localization, freeing clinicians from the burden of manual contouring while preserving the fidelity required for oncologic decision-making and only requiring minimal human interaction.

# 6    Tables

**Table 1.** Dataset splits and characteristics used during model training, validation and evalutation steps.

| Phase | Dataset | Study | Kind | Split | # Volumes | # Patients |
|---|---|---|---|---|---|---|
| **Training** | A | IRBd20-091 | Retrspective | Train | 50 | 50 |
|  |  |  |  | Validation | 10 | 10 |
| **Tetsing** | Bv | NL80634.031.22 | Retrospective | Test | 20 | 20 |
|  |  |  | Prospective | Test | 5 | 5 |

**Table 2.** Results for evaluation metrics over the retrospective test set (Sec. 2.1) for the three algorithms. Analysis for $Model_{full}$ was also performed exluding missed lesions to futher understand its performances.

|  | Semi-auto | | | $Model_{crop}$ | | | $Model_{full}$ | | | $Model_{full-no-zeros}$ | | |
|---|---|---|---|---|---|---|---|---|---|---|---|---|
|  | 25% | Med. | 75% | 25% | Med. | 75% | 25% | Med. | 75% | 25% | Med. | 75% |
| **Dice** | **0.76** | **0.83** | **0.87** | 0.57 | 0.74 | 0.81 | 0.00 | 0.02 | 0.57 | 0.47 | 0.60 | 0.69 |
| **Precision** | **0.91** | **0.93** | **0.94** | 0.61 | 0.76 | 0.86 | 0.00 | 0.75 | 0.91 | 0.83 | 0.91 | 0.96 |
| **Recall** | 0.65 | 0.74 | 0.81 | **0.66** | **0.80** | **0.89** | 0.00 | 0.01 | 0.42 | 0.32 | 0.47 | 0.64 |
| **VRD** | **0.13** | **0.21** | **0.29** | 0.11 | 0.27 | 0.54 | 0.50 | 0.86 | 1.00 | 0.27 | 0.43 | 0.65 |
| **HD** | 5.60 | 14.74 | 18.72 | 10.31 | 17.62 | 28.79 | 17.30 | 29.79 | 38.69 | 17.44 | 29.63 | 37.40 |
| **Time** | 201.29 | 243.70 | 320.15 | **49.75** | **54.50** | **61.25** | 58.77 | 65.36 | 71.69 | 58.90 | 65.44 | 71.21 |

# 7 Figures

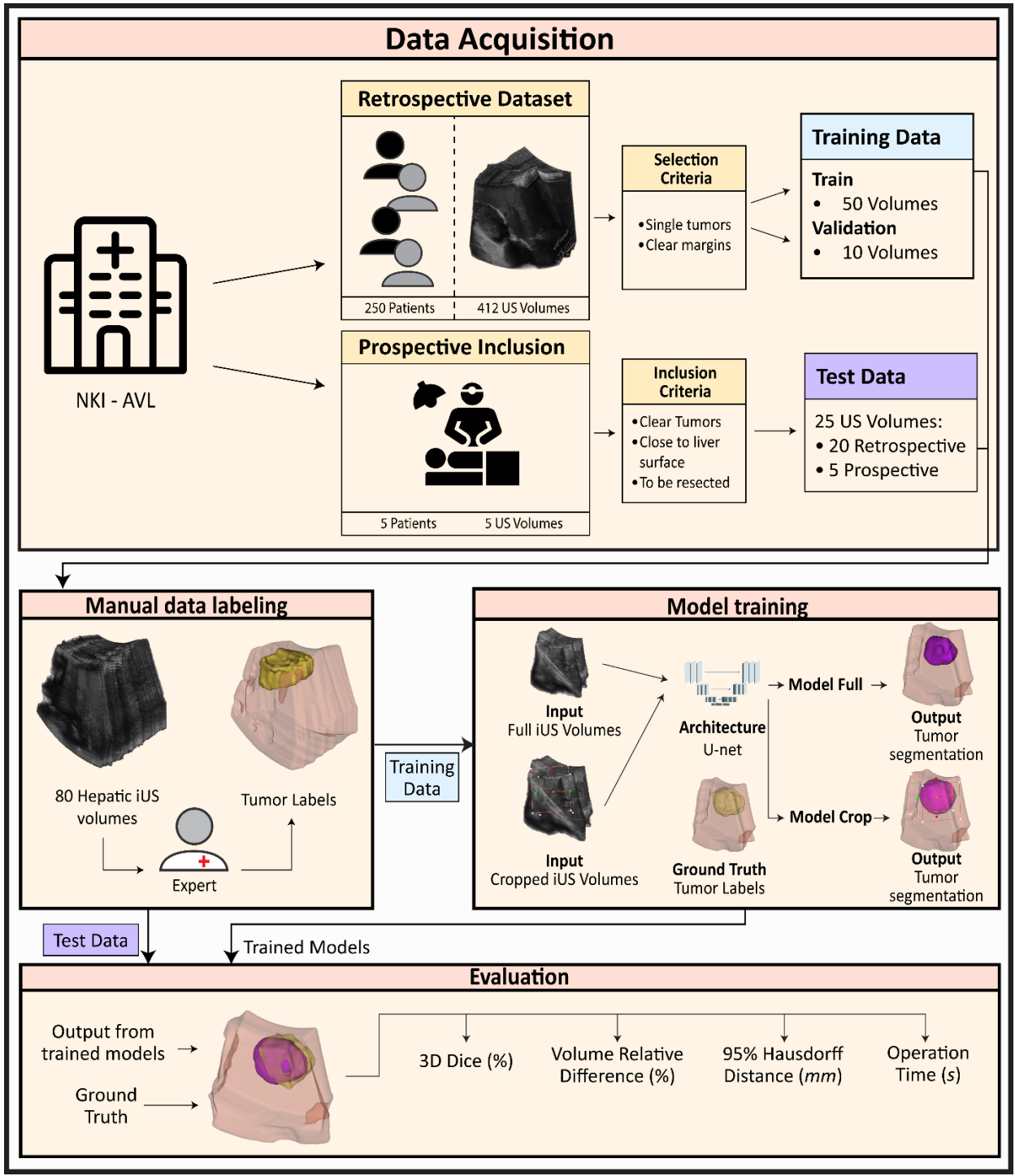

**Figure 1.** Structure of the presented study, from patient inclusion, to clinical workflow and evaluation.

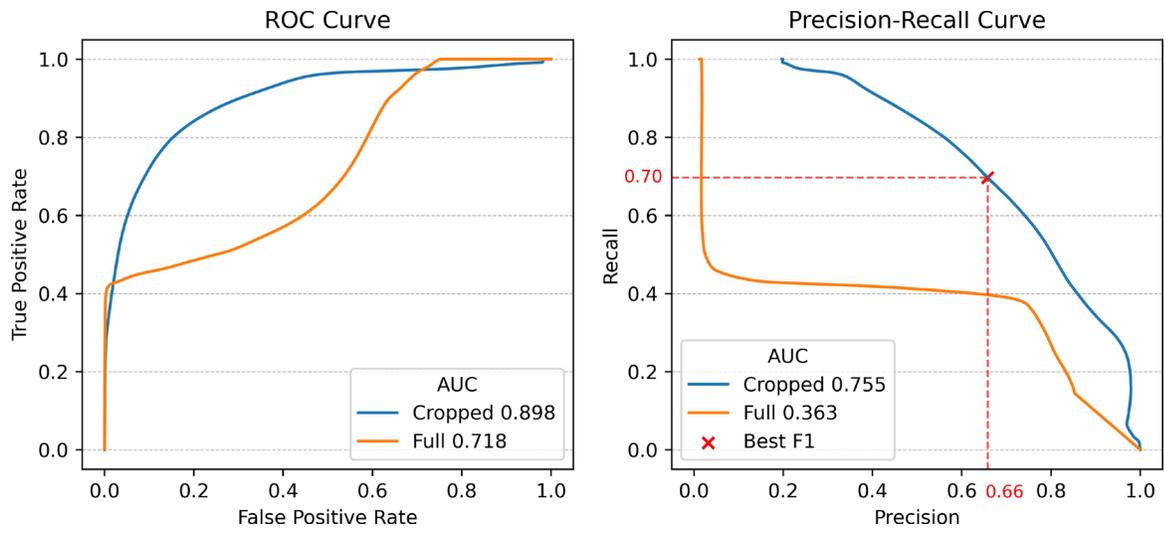

**Figure 2.** ROC and Precision Recall curve with selected thresholds

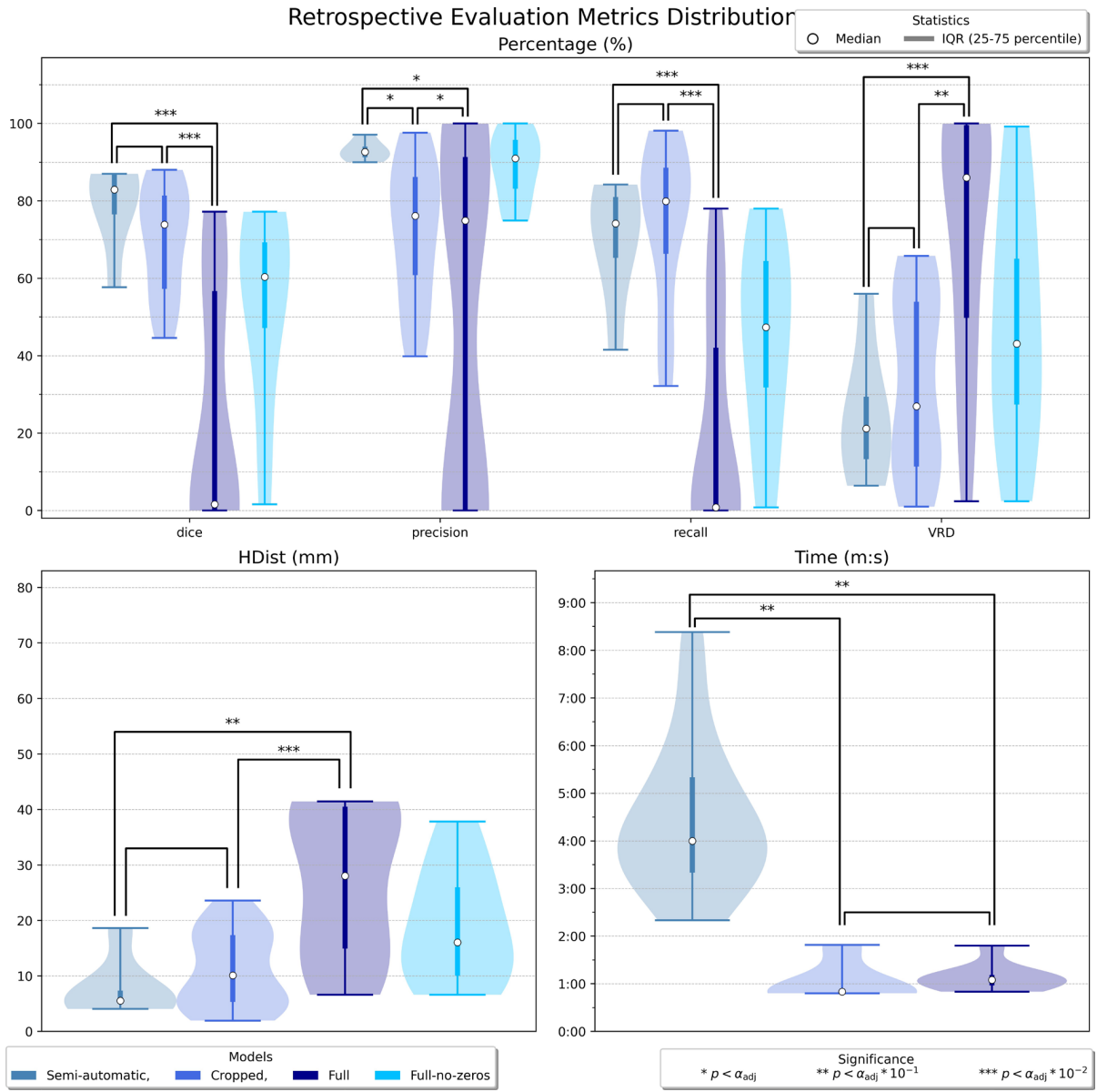

**Figure 3.** Results of the presented models over set $B_{retro}$ (Sec. 2.1)

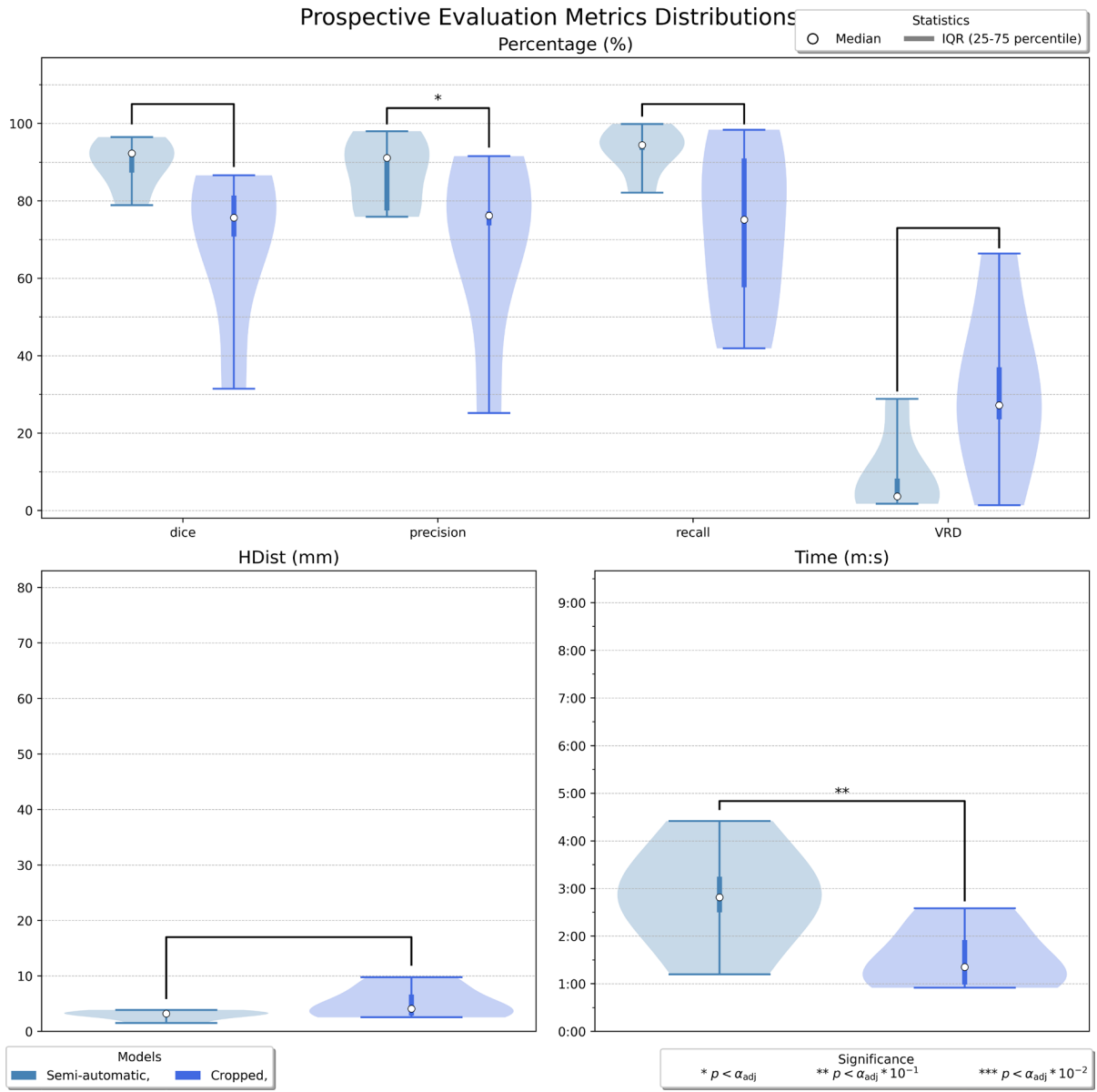

**Figure 4.** Results of the presented models over set $B_{pro}$ (Sec. 2.1). Violin plots showing segmentation performance and execution time for the prospective dataset. Accuracy remained comparable between the cropped and semi-automatic methods, while execution time slightly increased for the cropped model and decreased for the semi-automatic approach. Statistical significance followed the same trends as in the retrospective evaluation.

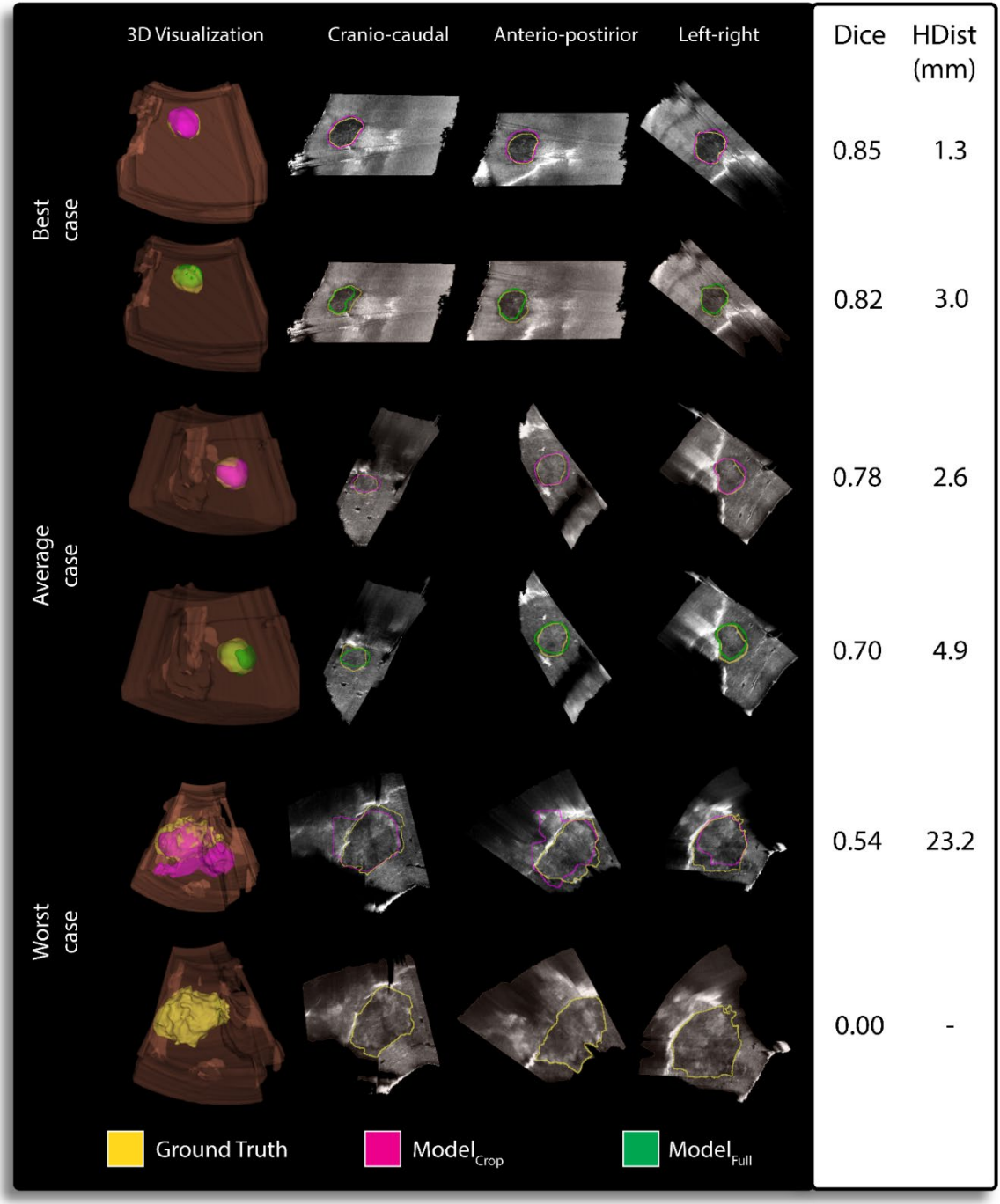

**Figure 5.** Qualitative analysis of 3 cases from the Test set (Tab. 1), representing a best, an average and a worst scenario. Ground truth delineations are in yellow, in magenta those produced by $Model_{crop}$ and in green those from $Model_{full}$. In the best case, both model produced delineations similar to the ground truth. In the average case, the models still manage to delineate the lesions with good accuracy, with lower accuracy than in the best case. In the worst case scenario, $Model_{full}$ does not detect the lesion, and consequently no segmentation is produced. $Model_{crop}$ still succeeds at segmenting the majority of the tumor, including false-positive areas in the segmentation which result in lower scores for Dice and HDist.